\documentclass[manuscript]{acmart}

\AtBeginDocument{%
  }

\copyrightyear{2022}
\acmYear{2022}
\setcopyright{acmcopyright}\acmConference[GoodIT'22]{Conference on Information Technology for Social Good}{September 7--9, 2022}{Limassol, Cyprus}
\acmBooktitle{Conference on Information Technology for Social Good (GoodIT'22), September 7--9, 2022, Limassol, Cyprus}
\acmPrice{15.00}
\acmDOI{10.1145/3524458.3547264}
\acmISBN{978-1-4503-9284-6/22/09}

\usepackage{xcolor}
\newcommand{\KK}[1]{{\color{red} (\underline{KK:} {#1})}}

\usepackage{multirow}
\usepackage{balance}
\usepackage{chato-notes}
\begin{document}

\title{LibertyMFD: A Lexicon to Assess the Moral Foundation of Liberty.}



\author{Oscar Araque}
\email{o.araque@upm.es}
\orcid{0000-0003-3224-0001}
\affiliation{%
  \institution{Universidad Polit\'ecnica de Madrid, ETSI Telecomunicaci\'on}
  \city{Madrid}
  \country{Spain}
}

\author{Lorenzo Gatti}
\email{l.gatti@utwente.nl}
\orcid{0000-0003-2422-5055}
\affiliation{%
  \institution{University of Twente}
  \city{Enschede}
  \country{The Netherlands}}
\email{}

\author{Kyriaki Kalimeri}
\email{kyriaki.kalimeri@isi.it}
\orcid{0000-0001-8068-5916}
\affiliation{%
  \institution{ISI Foundation}
  \city{Turin}
  \country{Italy}}

\renewcommand{\shortauthors}{Araque, et al.}

\begin{abstract}
Quantifying the moral narratives expressed in the user-generated text, news, or public discourses is fundamental for understanding individuals' concerns and viewpoints and preventing violent protests and social polarisation.
The Moral Foundation Theory (MFT) was developed to operationalise morality in a five-dimensional scale system.
Recent developments of the theory urged for the introduction of a new foundation, the \textit{Liberty Foundation}.
Being only recently added to the theory, there are no available linguistic resources to assess whether liberty is present in text corpora. 
Given its importance to current social issues such as the vaccination debate, we propose two data-driven approaches, deriving two candidate lexicons generated based on aligned documents from online news sources with different worldviews.
After extensive experimentation, we contribute to the research community a novel lexicon that assesses the liberty moral foundation in the way individuals with contrasting viewpoints express themselves through written text.
The \textit{LibertyMFD} dictionary can be a valuable tool for policymakers to understand diverse viewpoints on controversial social issues such as vaccination, abortion, or even uprisings, as they happen and on a large scale.

\end{abstract}

\maketitle

\section{Introduction}

Moral values are fundamental to our decision-making process on everyday matters. When taking a stance on a social issue, for instance, global warming or vaccine adherence, we consult - consciously or unconsciously - our moral system of values. 
Extracting and analysing moral content from user-generated text or public discourse, in general, is critical to understanding the decision-making process of individuals while getting a large scale perspective of evolving narratives~\cite{rao2020political}.
The Moral Foundations Theory (MFT) was created to explain morality across cultures \cite{Haidt2004}. The theory initially proposed five foundations, namely \textit{care, fairness, loyalty, authority}, and \textit{sanctity}, while more recently, the theory was enhanced with a new sixth dimension: \textit{liberty}. 

The MFT theoretical framework presents libertarians with a unique moral-psychological profile, endorsing the principle of liberty as an end and devaluing many of the moral concerns typically endorsed by conservatives.
Libertarianism is a political philosophy and movement that upholds liberty as a core principle~\cite{barnett2014structure} and expresses the extreme side of the liberty moral foundation. Analysing the psychological dispositions of libertarians, Iyer et al.~\cite{iyer2012understanding}, found that libertarians are consistently less concerned about individual-level concerns such as harm, benevolence, and altruism. They are also much less concerned with group-level moral issues, such as conformity, loyalty, and tradition, typically associated with conservative morality. Libertarians' cognitive style depends less on emotion and more on reason than conservatives. 
The liberty/oppression foundation, as described in the MFT, ``deals with the domination and coercion by the more powerful upon the less so'' \cite{haidt2012righteous}.
 
The MFT is broadly adopted in the computational social science
field since it defines a clear taxonomy of values together with a lexicon, the Moral Foundations Dictionary  (MFD)~\cite{graham2009liberals}, an essential resource for natural language processing applications.
The MFD creators highlight the difficulty of creating such a resource since linguistic, cultural, and historical contexts reflect on language usage. Among the most significant limitations of the original MFD, we have: (i) a limited amount of lemmas and stem of words; (ii) ``radical'' lemmas rarely used in everyday language, for instance, ``homologous'' and ``apostasy''; and (iii) an association with a moral bi-polar scale, so-called vice and virtue, but without any numeric indication of weight or importance to them.
(iv) the \textit{liberty} foundation is not considered due to its very recent addition to the main theory.
To address these shortcomings, other lexicons were developed; the most broadly used are the  MoralStrength lexicon~\cite{araque2020moralstrength} and the extended Moral Foundations Dictionary (eMFD)~\cite{hopp2021extended}.
The eMFD continues along with the original MFD lexicon, treating each word related to a moral as either one of two opposites, ``vice'' and ``virtue''. This approach leads to contradictory findings when both a vice and virtue are present in the same text, making it harder to assign a clear polarity to it. Following different approaches, both expand the number of lemmas per foundation with more commonly used terms.
MoralStrength on the other hand, inspired by the literature in sentiment analysis,  provides a notion of ``moral valence'' that ranges along the entire spectrum of each foundation from vice to virtue, providing more fine-grained and nuanced results.
Despite addressing several of the most critical shortcomings of the original lexicon, these resources do not include the \textit{liberty} foundation.

Here we lay the groundwork for a linguistic resource that can detect the presence of the \textit{liberty} moral dimension in people's narratives. 
Since this foundation was a later addition to the MFT theory, there are no initial linguistic indicators that can be used as seed words for further expanding the lexicon, a pre-requisite for the eMFD and MoralStrength approaches. 
To overcome this issue, we gathered data from news sources on most political ideologies and leanings in the USA political scene. More specifically, we obtained data from AllSides\footnote{\url{https://www.allsides.com/unbiased-balanced-news}}, an American news aggregator that presents different versions of similar news stories from sources of the political right, left, and centre \cite{AllSides} to help readers break through their filter bubble, as well as the libertarian magazine Reason\footnote{\url{https://reason.com}}. 
Seen through the lenses of our theoretical framework, the Moral Foundations Theory, Reason prioritises the moral value of liberty with respect to the news outlets from other political views.

We employ two data-driven approaches, the first based on word embedding similarity and the second on compositional semantics.  
To decide on the most efficient approach, we compare the two models qualitatively and quantitatively via two separate classification tasks. Firstly, we employ the obtained lexicons to predict whether an unseen document originated from AllSides or Reason datasets. Secondly, in an unsupervised manner, we infer the presence of the moral value of liberty in Facebook comments on Pages regarding vaccination.
We conclude that the lexicon generation based on the compositional semantics approach was overall the most efficient one, and hence the one we propose as the final \textit{LibertyMFD} lexicon.

In the age of social media, social crisis, disasters, and epidemics, drive not
only affliction in the physical world, but also prompt a deluge of opinions, (mis)information, and
advice to billions of internet users~\cite{mejova2020covid}.
Here, we contribute to the research and policymaking communities with a resource that facilitates the understanding of the liberty and moral value of user-generated communications on controversial phenomena timely and at large scale.

\section{Theoretical Background and Related Literature}

The original Moral Foundations Dictionary (MFD)~\cite{graham2009liberals} consists of a collection of lemmas assembled by experts and was typically used together with the Linguistic Inquiry and Word Count (LIWC) software~\cite{Tausczik2010} to estimate moral traits and to investigate differences in moral concerns between different cultural groups.
Garten et al.~\cite{Garten2018} proposed the Distributed Dictionary Representations (DDR) method based on psychological dictionaries and semantic similarity to quantify the presence of moral sentiment around a given topic. 
Later on, the authors extended the method, incorporating demographic embeddings into the language representations~\cite{Garten2019}.

In an attempt to address several of the limitations of the MFD, Araque et al.~\cite{araque2020moralstrength} proposed a data-driven generated lexicon the MoralStrength, which expanded the original MFD employing the WordNet synsets and crowdsourced annotations. Different from the MFD, where each foundation is considered a bipolar of ``virtue'' and ``vice'', MoralStrength treats each foundation as a continuum,  assigning a numeric value of moral valence to each lemma that indicates the weight with which the lemma is expressing the specific value. 
Hopp et al.~\cite{hopp2021extended} developed the extended Moral Foundations Dictionary (eMFD), a lexicon which expanded the MFD based on crowdsourced annotations. Each lemma in eMFD is assigned a continuously weighted vector that expresses the probability that the lemma belongs to any of the five moral foundations. 

Notably, none of the above lexicons though included the liberty moral foundation. 
A first attempt to derive a lexicon from assessing the presence of liberty in the text was presented by Araque et al.~\cite{araque2021liberty}. They considered pairs of Wikipedia Pages\footnote{\href{https://www.wikipedia.org}{Link to the Wikipedia Project.}} and their Conservapedia\footnote{\href{https://www.conservapedia.com/Main_Page}{Link to the Conservapedia Project.}} counterparts as a natural experiment. They created a series of word embeddings which were then compared through cosine similarity to a set of seed words defined by experts to generate a lexicon. 
Although innovative, their design comes with the obvious conceptual limitations of considering the Wikipedia project as expressing a strongly libertarian position and initiating the embeddings with a list of manually selected seed words from expert annotators. 
Here, we overcome these limitations by considering news outlets with a clear political orientation provided by experts in the field\footnote{\href{https://www.allsides.com/media-bias/media-bias-chart}{Media Bias Chart provided by AllSides}}. At the same time, our initial seed words emerge in a data-driven way from the actual linguistic behavioural patterns employed by the two sides.

\section{Materials and Methods}\label{sec:methods}
Our initial assumption is that libertarians are highly sensitive to the moral value of liberty (and lack thereof). Hence, their communication will contain more words related to the liberty foundation than people self-identifying with a different part of the political spectrum.
We propose two entirely data-driven models for assessing the liberty foundation aiming to obtain a final \textit{LibertyMFD} lexicon.
The first model (Sect.~\ref{subsec:method-word-embedding}) is based on word embeddings, and the second one (Sect.~\ref{subsec:method-cs}) on computational semantics.

\subsection{Data Collection}
To assess the linguistic patterns of the entire political spectrum of the U.S., we obtained data from the AllSides and Reason platforms. Moreover, we included the only available resource with annotations regarding the liberty moral foundation          \cite{Prado2022}.

\subsubsection{News Dataset.}
The \textit{Reason} is an American libertarian monthly magazine published by the Reason Foundation. To obtain a representative sample of data expressing the libertarian point of view on multiple societal issues, we downloaded 14,319 articles from the website, ranging from June 1968 to June 2021\footnote{\href{https://reason.com/about/}{Reason.com. Retrieved May 27th, 2022}. Data downloaded according to the \href{https://reason.com/terms-of-use/}{Terms of Use} from \url{https://reason.com/archives/}.}.
About 50 documents were discarded, as they contained no text.
Each article (averaging 1235 words in length ) consists of a title, publication date, and the magazine section where it appears. The articles published after 1994 usually contain a summary, one or more keywords and the author's name.

\textit{AllSides} is an American company that assesses the political bias of prominent media outlets, aligning similar news stories from sources of the political right, left, and centre in a mission to show readers news outside their filter bubble. 
We collected 5,584 ``news roundups''\footnote{\url{https://www.allsides.com/story/admin}} published between June 2013 and April 2021. These consist of a short description of the news item, the headline and the first paragraphs (71 words on average, including headline and description, calculated on all 11,379 individual articles downloaded) of the original publication and its source. The AllSides media bias chart\footnote{\href{https://www.allsides.com/news-source/reason}{Classification of Reason magazine.}}
classifies Reason as the only libertarian-leaning source; however, they opted to assign articles originating from Reason magazine with the ``right-leaning'' political side. Here, we excluded all 67 articles from Reason included in the AllSides dataset.

To ensure the stability and generalisability of the approach, we split this dataset in two (Table~\ref{tab:data-train-test}). The training set, consisting of 80\% of the whole \textit{News} dataset, is used for the generation of the \textit{liberty} lexicons.
 The left out set, i.e. the remaining 20\% articles, will only be used for testing the quality of the generated lexicons.

\subsubsection{Vaccine Dataset.} To evaluate the obtained lexicon, we also employ data from Facebook Pages with a clear stance about vaccination~\cite{Prado2022}. The data include anonymised posts and comments from approximately 200 Facebook Pages collected via the Facebook API from January 2012-June 2019. 
Overall, there are  $607,105$ comments and posts from vaccines supporting and opposing sides. Approximately 1,500 comments were randomly selected and annotated manually as to whether the liberty moral foundation is present or not in the text snippet and whether it is associated with the ``virtue'' (liberty) or ``vice'' (oppression) polarity of the dimension.
It is worth mentioning that this dataset, although limited, is the only available resource containing annotations about the moral foundation of liberty. In contrast, the most commonly used benchmark dataset for analysing the moral foundations~\cite{hoover2020moral} does not include the liberty/oppression foundation.

\begin{table}[ht]
\caption{Overview of the datasets employed in this study. The \textit{News} dataset was used for the generation of the lexicons (Training Set), while for the evaluation of the performance, only the Left-out Set from the \textit{News} dataset was employed. The \textit{Vaccine} dataset was used for the evaluation of the lexicons. For each set, the total number of instances is provided.}
\begin{tabular}{llc}
    \renewcommand{\arraystretch}{1.2}
 \textbf{Dataset} & \textbf{Label} & \textbf{No. of instances (\%)} \\
\toprule
\multirow{2}{*}{\textit{News}  (Train Set  - Used to generate the lexicons)} & Liberty/Oppression (Reason) & 11,470  (56\%)  \\
& Other (AllSides) & 9,062 (44\%) \\
\midrule
\multirow{2}{*}{\textit{News}  (Left-out Set  - Used to evaluate the lexicons)} & Liberty/Oppression (Reason) & 2,849 (56\%) \\
& Other (AllSides) & 2,284 (44\%) \\
\midrule
\multirow{2}{*}{\textit{Vaccine} (Used to evaluate the lexicons)} & Liberty/Oppression (annotations) & 178 (11\%) \\
& Other &  1398 (89\%)  \\
\bottomrule
\end{tabular}
\label{tab:data-train-test}
\end{table}


\subsection{Word Embedding Similarity}
\label{subsec:method-word-embedding}

Inspired by the work of Turney et al.~\cite{turney2003measuring}, the first approach to generate the \textit{LibertyMFD} is based on word embedding similarity among vectors of both AllSides and Reason. 
This approach requires an initial set of seed words that represent the domains we aim at distinguishing, in our case, the liberty/oppression moral dimension from the \textit{others}, namely all remaining moral foundations (such as authority, purity) and non-moral text.
Due to the lack of any apriori linguistic information, in previous work \cite{araque2021liberty} experts manually defined the list of seed words; however, slight variations in the seed word list may generate very different final lexicons. 

Here, to avoid the shortcomings of an arbitrary selection, we obtain the set of seed words in a data-driven way, estimating the \textit{frequency shifts}~\cite{gallagher2021generalized} of the lemmas between the AllSides and Reason datasets.
Let the relative frequency of a word $w$ in a set of documents $D$ be:

$$ p_w^{(D)} = \frac{f_w^{(D)}}{\sum_{w' \in W^{(D)}} f_{w'}^{(D)}} $$

\noindent where $w' \in W^{(d)}$ are the words in vocabulary set $W^{(D)}$ except for $w$.
We compute the frequency shift in the relative frequency per word $w$ between two different sets of documents:

$$ \delta p_w  = p_w^{(2)} - p_w^{(1)} $$

\noindent The corresponding seed word lists emerge as salient differences in word frequency shifts.
We apply a minimum frequency threshold at 100, under which a lemma is filtered out of the seed word list.
Then, we compute each word's respective vector employing the word2vec algorithm~\cite{le2014distributed} with a standard parameter setting and a vector dimension equal to 100. 
The lexicon is then generated by estimating the cosine similarity between the word vectors obtained with the emerged seed words.
If $S_C$ is the set of seed words for the ``others'' orientation, and $S_L$ the seed words for the ``liberty'' direction, the moral polarity of a word $w_i$ is computed from the documents as

$$\sum_{w_j \in S_L} \text{sim}(w_i, w_j) - \sum_{w_k \in S_C} \text{sim}(w_i, w_k) $$
where \textit{sim} is the cosine similarity estimated by the word embedding model.
The polarity is a positive number if $w_i$ is related to the ``liberty/oppression" seed words and a negative one if the relation occurs towards the ``other'' seed words.
For simplicity, we will refer to this model as the \textit{WE model}.

\subsection{Compositional Semantics}
\label{subsec:method-cs}
The second approach is based on the \textit{Compositional Semantics} (CS) method, which has been extensively used to generate emotion lexicons~\cite{staiano2014depeche,depechemoodpp}.
We adapt the domain of application to the liberty moral values; in this case, we consider the documents of Reason as discussing liberty/oppression more frequently than the documents from AllSides. 
The CS method projects the moral values from a document to its words, assuming that each word is associated with the moral value expressed in the documents where the word appears more frequently.

Let us consider a document-by-moral matrix $M_{DM}$ which contains the distribution of the liberty foundation across the Reason and AllSides datasets.
We compute a word-by-document matrix $M_{WD}$ which contains the number of occurrences for each word in the vocabulary in a given document, normalised with the number of words per document.
To obtain the word-by-moral matrix, we perform the following multiplication:

$$ M_{WM} = M_{WD} \cdot M_{DM} $$

\noindent This allows us to merge words with moral values by summing the products of the weight of a word with the weight of the moral values in each document.
Finally, we normalise the lexicon scores (column-wise), remove over-representation issues, and scale each lemma (row-wise) to sum up to one. 
Previous lexicon validations showed that this is an adequate normalisation approach~\cite{depechemoodpp}.
For simplicity, we will refer to this model as the \textit{CS model}.

\subsection{Evaluation Design}
\label{sec:evaluation}

\subsubsection{Lexicon Overlap Score and Coverage.}

We employed a series of qualitative and quantitative methods to gauge the differences between the two generated lexicons, namely the WE and the CS lexicons.
For each lexicon, we estimate the average coverage per dataset as a percentage of the total number of lemmas in the dataset that also exist in the respective lexicon.
Then, we assess the statistical properties of the liberty score distributions obtained from two lexicons and provide 
excerpts covering both polarities of each lexicon.
We also provide the lexicon overlap score (LOS) as defined in \cite{welter2020quantitative}.
The simple version of the lexicons overlaps score considers the size of both
lexicons and the words they have in common. Since the simple LOS does not consider the general direction of the polarity, we also report the binary LOS.

\subsubsection{Unsupervised Document Categorisation.} 
We perform an unsupervised evaluation of the WE and CS lexicons on the \textit{Vaccine} dataset in line with the evaluation approach proposed for the comparison of polarity lexicons in sentiment analysis tasks~\cite{welter2020quantitative}.
For each document, we average the liberty scores assigned to the lemmas in the lexicon under evaluation. 
In our case, a document is a comment on a Facebook Page post.
Hence, we obtain two numeric scores for each document, describing how much the document expresses the value of liberty according to the WE and the CS lexicons, respectively. 
We then binarise the scores according to the median value of the respective lexicon, deciding for each document whether it expresses  ``liberty/oppression" moral values or not.
Negation correction was not applied, as foundation polarities do not directly translate as opposites (e.g., ``not libertarian'' is not the same as ``oppressive'').
Here, in a very straightforward way, we assess the extent to which the two lexicons assign a common or a diverse label to an unseen document.

\subsubsection{Supervised Classification of Liberty.}
To evaluate the performance of the two lexicons, we designed two supervised classification tasks. 
In the first one, a classifier predicts whether an unseen document from the \textit{News} dataset was published by Reason (i.e., expresses a libertarian position) or from AllSides. The classifier is trained (and evaluated, using cross-validation) on the left-out set of articles that were not considered for generating the lexicons.
Similarly, for the \textit{Vaccine} dataset, we aim to accurately infer the labels assigned by the expert annotators on whether a post or a comment contains the liberty foundation (regardless of whether it is expressed as virtue or vice) or not. This second experiment aims at determining to what extent the lexicons help identify the presence of liberty in new datasets and domains, going beyond the limits of our na\"ive unsupervised approach. 

We trained an SVM classifier \cite{fan2008liblinear} with a linear kernel for each one of the above tasks and each lexicon. 
The feature vectors are constructed as follows; each document is represented by a vector of equal size to the lexicon. For those lemmas in the document present in the lexicon, the vector contains the respective polarity score, otherwise zero.
Since this type of representation dramatically simplifies the linguistic information present in the document, we enhance the classification design with two more experiments. We extend each vector representation with the ``statistical summary'' functions, namely the average, maximum, median, variance, and a peak-to-peak score of the lexicon values of that document. This offers the learning models a more complete view of the text.

As a baseline model, we train an SVM classifier using a unigram representation with a standard vocabulary size of 5,000 tokens. We report the macro-averaged F-score with 10-fold cross-validation for all experiments. We also test and report the performance of the ``statistical summary'' used in isolation.

\section{Results and Discussion}

This study provides a thorough comparison of the two lexicons employing both supervised and unsupervised approaches.

\subsection{Lexicon Overlap Score and Coverage.}

\begin{figure}[htb]
    \centering
    \includegraphics[width=0.7\linewidth]{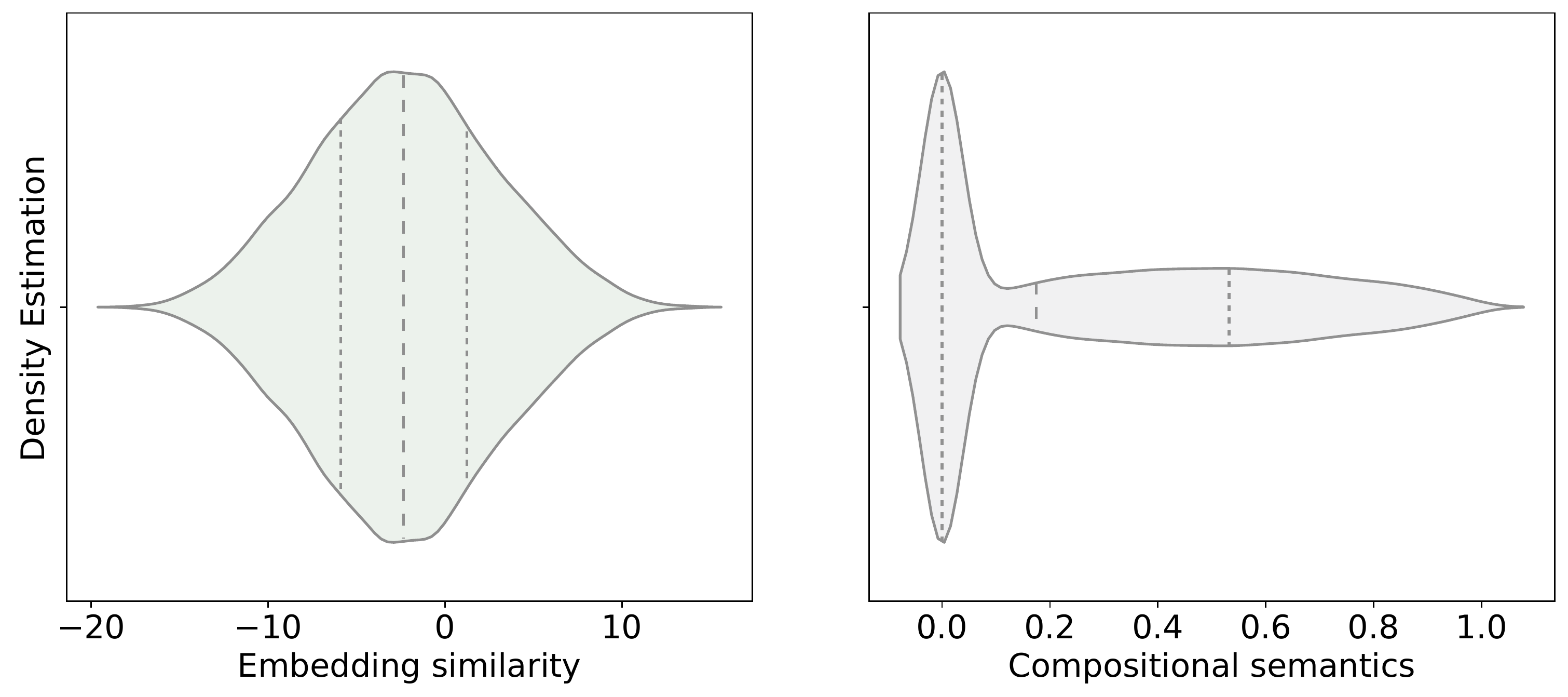}
    \caption{Distributions of the generated lexical resources, the word embedding similarity, WE model, on the left hand side and the Compositional Semantics, CS model on the right hand side. Dashed lines indicate the quartiles of the distributions.}
    \label{fig:lexicons-stats}
\end{figure}

Figure~\ref{fig:lexicons-stats} depicts the distribution of moral scores for the two generated resources. The score distributions from the two lexicons are statistically different ($p<0.001$), evident from both the spread and shape. 
The WE model has a unimodal distribution with a skewed average towards the negative values. The liberty values of the CS model instead have a right-skewed bimodal distribution with an imposing first mode and a long tail.

Another interesting characteristic is that the average value of the WE lexicon distribution is skewed to the negative values, while the first mode of the CS lexicon is 0. Intuitively, this is reasonable, as it indicates that most of the lemmas are not indicative of the \textit{liberty} foundation.
The obtained distributions follow patterns found in previous lexicons; in fact, the normal distribution of the WE lexicon is similar to that of SentiWordNet~\cite{baccianella2010sentiwordnet}, while the CS lexicon has a similar distribution to the NRC Emotion lexicon~\cite{mohammad2016hashtags,araque2020radicalization}.

Checking the coverage of the two models, we observe that the coverage of the CS lexicon is much higher across all datasets, as reported in Table~\ref{tab:lexicons-coverage}.
The simple LOS states to what extent the lexicons share a common vocabulary 77.2\%, while the binary LOS is 39.5\%. At the same time, if we consider the lemmas with a frequency of appearance above the average for both lexicons, we obtain 93.5\% shared lemmas, with a binary LOS of 44.4\%.
The above finding shows that the two lexicons are more in accordance with polarity scores for the most popular lemmas.

Moreover, the two methods capture different patterns of language usage related to \textit{liberty}. As can be seen in Table \ref{tab:lexicons-lemmas}, ``expense'' and ``distribution'' are both highly related to liberty/oppression. Given that the lexicons are generated starting from news, these terms likely refer to economic spending and usage of taxes, but their relative order is different in the two resources.

\begin{table}[ht]
\renewcommand{\arraystretch}{1.2}

\caption{Average coverage of the two lexicons over all datasets. Coverage is measured with the number of words from each document that appear in each lexicon.}
\begin{tabular}{l c c}

\textbf{Dataset} & \textbf{WE lexicon coverage} & \textbf{CS lexicon coverage} \\
\toprule
News Train Set & 82.0 & 96.8 \\
News Left-out Set & 82.0 & 96.6 \\
Vaccine & 77.4 & 90.9 \\
\bottomrule\end{tabular}

\label{tab:lexicons-coverage}
\end{table}

\begin{table}[h]
\renewcommand{\arraystretch}{1.2}
\centering
\caption{Example lemmas from the WE and CS lexicons. We indicatively present some of the lemmas with the highest and lowest polarity score for the two lexicons.}
\begin{tabular}{p{0.05\linewidth}p{0.1\linewidth}p{0.76\linewidth}}

\textbf{Model} & \textbf{Liberty Polarity}& \textbf{Lemmas} \\ 
\toprule
WE & High & distribution,knowledge,wealth,control,function,productive,expense,values,producing, skills\\
WE & Low & Georgia,senator,Harris,Iowa,governor,congressman,Ohio,impeachment,scandal\\
CS & High &corporations,coercion,graduate,develop,incomes,budgets,expense,scientists,advice\\
CS & Low &  financial,Americans,enemy,responsible,suspect,citizens,racial,bias,suburban,refuse\\
\bottomrule
\end{tabular}

\label{tab:lexicons-lemmas}
\end{table}

\subsection{Unsupervised Document Categorisation.}
We further evaluate the WE  and CS lexicon differences, respectively, in an unsupervised classification task. Here, we aim to infer whether a document, from the \textit{Vaccine} dataset, expresses a concept relevant to the moral foundation of liberty.
We evaluate the outcome comparing against the manual annotations provided by experts.
Given the limited linguistic coverage observed already in Table~\ref{tab:lexicons-coverage}, the performance is expected to be relatively low.
Nonetheless, this evaluation offers the first indication of the performance of the two models.
An average macro F-score of 45.32 for the CS lexicon, against 23.46 for the WE one, gives a clear benefit to the CS model. 

\subsection{Supervised Classification of Liberty.}

We evaluated the obtained lexicons as per their predictive power in two independent text classification tasks on both the \textit{News} and the \textit{Vaccine} datasets.
As described in Sect.~\ref{sec:evaluation}, we compare the classification performance of the two proposed lexicon generation methods, including a baseline for reference.

Table~\ref{tab:sup-evaluation} reports the average macro F-score and the standard deviation obtained on a 10-fold cross-validation setup.
Worth mentioning is the big difference in performance between the two datasets; this should not surprise, as the \textit{Vaccine} dataset is significantly smaller, is high class imbalanced, and 
is focused on a domain that is very different from that of news.
The baseline model reaches a high performance in both datasets;
in the case of the \textit{News} dataset, both WE and CS generated lexicons that improve the performance over the baseline. For the \textit{Vaccine} dataset instead, only the CS models provides a slightly better average.
The difficulty of the WE model to adapt to a new dataset like the \textit{Vaccine}  may be related to the limited and fixed amount of information that can be encoded~\cite{lenci2018distributional}.

The ability of the CS model to generalise better in the \textit{Vaccine} dataset prediction, outperforming the baseline and the WE model, demonstrates the importance of having extended coverage.
Moreover, the long tail distribution of the moral scores of the CS model help distinguish more efficiently documents of a lesser-known domain not just in terms of vocabulary but also of writing style. 
Such ability can be parallelized with the human semantic ability to compose lexical meanings to form a potentially unlimited number of complex linguistic expressions. 

Interestingly, the introduction of basic statistical summary features alone is not informative of the liberty foundation; however, when added to the main feature vectors, it gives the classifier a complete view of the presence of the \textit{liberty} foundation in text.
To conclude, both the CS and WE model, with or without the statistical summary, can easily outperform a unigram baseline, showing the value of both lexicons in the domain of news. 
When considering a novel, unseen domain, our findings suggest that the compositional semantics approach (CS model) provides with a more efficient, domain-independent lexicon; hereafter we will refer to the output of the CS model as the \textit{LibertyMFD lexicon}.

\begin{table}[ht]
\renewcommand{\arraystretch}{1.2}

\caption{Macro averaged f-score in the \textit{News} left-out set and \textit{Vaccine} datasets. WE and CS models are the lexicons generated with the embedding similarity and compositional semantics approaches. In bold, we indicate  the best performing model for each dataset.}
\begin{tabular}{l c c}

\textbf{Model} & \textbf{News Dataset} & \textbf{Vaccine Dataset} \\
\toprule
Unigram (baseline) & 94.29 ($\pm 0.91$) & 76.07 ($\pm$ 5.13) \\
\midrule
WE model & 97.54 ($\pm$ 0.82) & 73.85 ($\pm$ 5.29)  \\
Stat. summary (WE) & 82.35 ($\pm$ 3.52) & 37.62 ($\pm$ 3.26) \\
WE model + Stat. summary & \textbf{98.42} ($\pm$ 0.44) & 74.21 ($\pm$ 5.03) \\
\midrule
CS model & 97.40 ($\pm$ 0.73) & 76.70 ($\pm$ 4.80) \\
Stat. summary (CS) & 82.60 ($\pm$ 1.76) & 44.41 ($\pm$ 3.32) \\
CS model + Stat. summary & 98.27 ($\pm 0.57$) & \textbf{77.51} ($\pm$ 5.08) \\
\bottomrule\end{tabular}
\label{tab:sup-evaluation}
\end{table}

\section{Conclusions}

The foundation of Liberty centres on individualism and independence and can have multiple facets ranging from social justice advocacy for immigrants to resistance to oppressive regulations with prosocial character~\cite{brennan2012libertarianism}. 
Literature on computational social science highlighted the importance of the liberty foundation in understanding peoples' fundamental rationale towards critical social issues such as attitudes towards vaccination~\cite{kalimeri2019human}, charitable giving~\cite{nilsson2020moral}, familial coercion and human rights~\cite{Clifford2015}.

Although the impact of the liberty foundation goes well beyond these topics, and despite the increasing body of research around the main five dimensions of the moral foundation's theory, there is still no linguistic resource assessing the liberty foundation.
Here, we address precisely this shortcoming.
Having as a cornerstone assumption that libertarian journals will express more intensively the value of liberty than the others, we considered the online news sources as a natural experiment to train and test our models. 

Words in the \textit{LibertyMFD} were selected after extensive experimentation on two different data-driven lexicon generation approaches. 
Further, we assessed the performance of the lexicon on a manually annotated dataset of user-generated text regarding the issue of vaccine hesitancy, one of the topics where liberty is known to be a critical factor~\cite{betti2021detecting}. 
Our findings showed that the lexicon generated based on the compositional semantics approach had a broader coverage even in an out of domain dataset. At the same time, the polarity scores assigned to each lemma appeared to be more predictive of the liberty moral foundation in unseen documents.  
As a future improvement, we aim to refine the lexicon polarity scores to achieve a more robust association with the two polarities of the liberty foundation.

The proposed resource is generated for the English language; although the framework is directly expandable to other languages it would require retraining of the model on local language datasets to efficiently grasp the linguistic particularities of each language.
We recommend practitioners who aim to utilise the resource in languages different than English simply via automatic text translation, to 
interpret the results with caution, since the moral expression of the words may differ across languages.


We contribute to the research and policymakers communities with an open-source lexicon that can be particularly helpful to understanding better moral judgments, dispositions, and attitudes formation from spontaneous digital data.
Given the penetration of social media in our communication and information ecosystem, 
such resources can be employed to analyse large scale user-generated texts nowcasting peoples' attitudes towards fast-evolving social phenomena with direct implications for the humanitarian sector.

\begin{acks}
  The work of Oscar Araque was supported by the European Union’s Horizon 2020 Research and Innovation Programme under project Participation (grant  agreement  no. 962547).
  KK acknowledges support from the Lagrange Project of the Institute for Scientific Interchange Foundation (ISI Foundation) funded by Fondazione Cassa di Risparmio di Torino (Fondazione CRT).
\end{acks}

\balance

\bibliographystyle{ACM-Reference-Format}
\bibliography{GoodIt}


\end{document}